\documentclass[10pt,twocolumn,letterpaper]{article}

\usepackage[pagenumbers]{cvpr} 

\usepackage{graphicx}
\usepackage{amsmath}
\usepackage{amssymb}
\usepackage{booktabs}
\usepackage{resizegather}

\usepackage[accsupp]{axessibility}  

\usepackage{multirow}

\newlength\savewidth

\makeatletter\renewcommand\paragraph{\@startsection{paragraph}{4}{\z@}
  {.5em \@plus1ex \@minus.2ex}{-.5em}{\normalfont\normalsize\bfseries}}\makeatother

\usepackage[capitalize]{cleveref}
\crefname{section}{Sec.}{Secs.}
\Crefname{section}{Section}{Sections}
\Crefname{table}{Table}{Tables}
\crefname{table}{Tab.}{Tabs.}

\def\cvprPaperID{3689} 
\def\confName{CVPR}
\def\confYear{2022}

\begin{document}

\newcommand{\methodname}{CLRNet}

\title{CLRNet: Cross Layer Refinement Network for Lane Detection}


\author{
Tu Zheng\textsuperscript{1,2$\ast$}~~~~Yifei Huang\textsuperscript{1,2$\ast$}~~~~Yang Liu\textsuperscript{1,2}~~~~Wenjian Tang\textsuperscript{1}~~~~Zheng Yang\textsuperscript{1}~~~~\\Deng Cai\textsuperscript{1,2}~~~~Xiaofei He\textsuperscript{1,2}~~~~ \\
\textsuperscript{1}Fabu \qquad
\textsuperscript{2}Zhejiang University \\
}

\maketitle


\begin{abstract}
    Lane is critical in the vision navigation system of the intelligent vehicle. Naturally, lane is a traffic sign with high-level semantics, whereas it owns the specific local pattern which needs detailed low-level features to localize accurately. Using different feature levels is of great importance for accurate lane detection, but it is still under-explored. In this work, we present Cross Layer Refinement Network~(CLRNet) aiming at fully utilizing both high-level and low-level features in lane detection. In particular, it first detects lanes with high-level semantic features then performs refinement based on low-level features. In this way, we can exploit more contextual information to detect lanes while leveraging local detailed lane features to improve localization accuracy. We present ROIGather to gather global context, which further enhances the feature representation of lanes. In addition to our novel network design, we introduce Line IoU loss which regresses the lane line as a whole unit to improve the localization accuracy. Experiments demonstrate that the proposed method greatly outperforms the state-of-the-art lane detection approaches. 
\end{abstract}
\let\thefootnote\relax\footnote{$^\ast$Equal contribution.}


\section{Introduction}
\label{sec:intro}
Lane detection is an important yet challenging task in computer vision, which requires the network to predict lanes in an image. Detecting lanes can benefit many applications, such as autonomous driving and the Advanced Driver Assistance System~(ADAS), which helps intelligent vehicles localize themselves better and drive safer.

\begin{figure}[t]
  \centering
  \begin{subfigure}[b]{0.49\linewidth}
    \includegraphics[width=\linewidth]{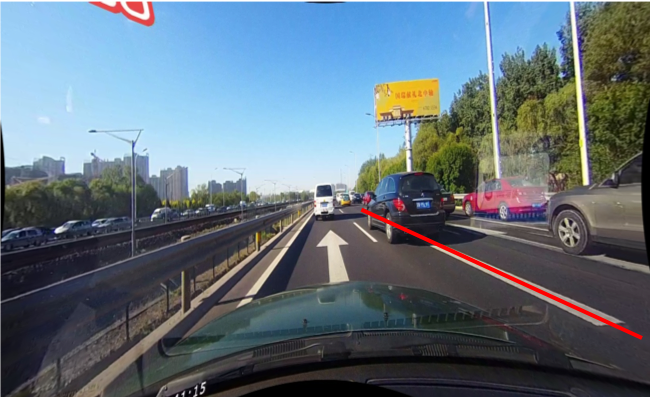}
    \caption{}
  \end{subfigure}
  \begin{subfigure}[b]{0.49\linewidth}
    \includegraphics[width=\linewidth]{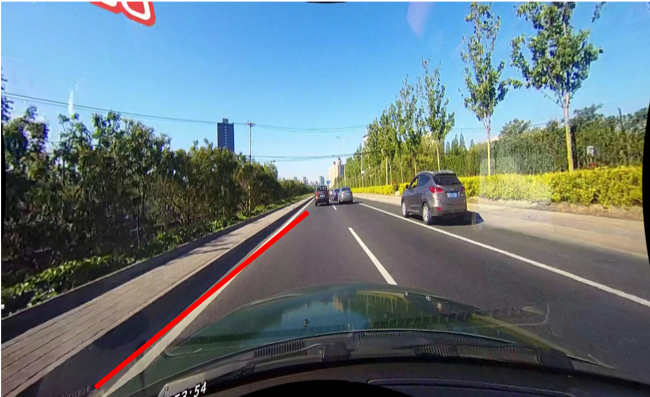}
    \caption{}
  \end{subfigure}
  \begin{subfigure}[b]{0.49\linewidth}
    \includegraphics[width=\linewidth]{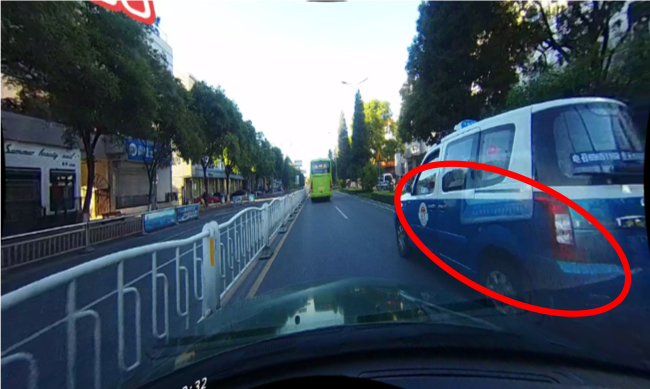}
    \caption{}
  \end{subfigure}
  \begin{subfigure}[b]{0.49\linewidth}
    \includegraphics[width=\linewidth]{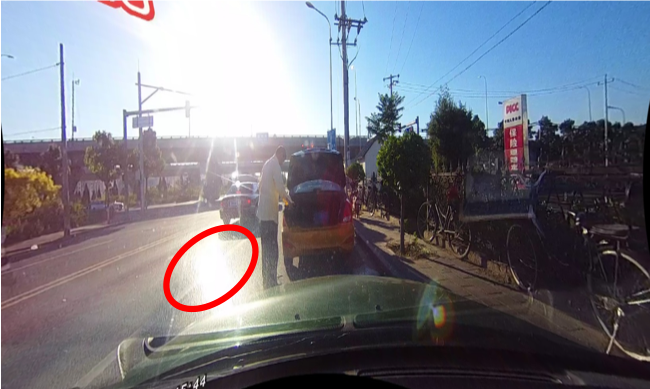}
    \caption{}
  \end{subfigure}
  \caption{Illustrations of hard cases for lane detection. (a) The detection result of low-level features. It mistakes landmark as lane due to losing global context. (b) The detection result of high-level features. It predicts inaccurate localization of the lane. (c) The case that lane is almost occupied by the car. (d) The case that lane is blurred by the extreme lighting condition.  }

  \label{fig:intro}
\end{figure}

Benefiting from the effective feature representation of CNN, many approaches~\cite{pan2018spatial, zheng2021resa, qin2020ultra} have obtained promising performance. However, there are still some challenges for detecting accurate lanes. 
Lane has high-level semantics, whereas it owns the specific local pattern which needs detailed low-level features to localize accurately. How to utilize different feature levels effectively in CNN remains a problem. As we can see in Fig.~\ref{fig:intro}(a), the landmark and lane line have different semantics, but they share the similar feature~(\emph{e.g.}, the long white line). It is hard to distinguish them without high-level semantics and global context. On the other hand, the locality is also essential since lane is long and thin with  the simple local pattern. We show the detection result of high-level features in Fig~\ref{fig:intro}(b), though the lane is detected, its location is not precise. Thus, the low-level and high-level information are complementary for accurate lane detection. 
Previous works either model local geometry of lanes and integrate them into global results~\cite{qu2021focus}  or construct a fully-connected layer with global features to predict lanes~\cite{qin2020ultra}.
These detectors have demonstrated the importance of local or global features for lane detection, but they don't take advantage of both features, yielding inaccurate detection performance.

Another common problem in lane detection is no visual evidence for the presence of lanes. As shown in Fig.~\ref{fig:intro}(c), the lane is occupied by the car while in Fig.~\ref{fig:intro}(d), the lane is hard to recognize due to the extreme lighting condition. In the literature, SCNN\cite{pan2018spatial} and RESA\cite{zheng2021resa} propose a message-passing mechanism to gather global context, but these methods perform pixel-wise prediction and don't take lane as a whole unit. Thus their performances lag behind many state-of-the-art detectors.


In this paper, we propose a new framework, Cross Layer Refinement Network~(\methodname), which fully utilizes low-level and high-level features for lane detection. Specifically, we first perform detection in high semantic features to coarsely localize lanes. Then, we perform refinement based on fine-detail features to get more precise locations. Progressively refining the location of lane and feature extraction leads to high accuracy detection results. To solve the problem of non-visual evidence of lane, we introduce ROIGather to capture more global contextual information by building the relation between the ROI lane feature and the whole feature map. Moreover, we define the IoU of lane lines and propose the Line IoU~(LIoU) loss to regress the lane as a whole unit and considerably improve the performance compared with standard loss, \emph{i.e.}, smooth-$l_1$ loss.

We demonstrate the effectiveness of our method on three lane detection benchmarks, \emph{i.e.}, CULane~\cite{pan2018spatial}, Tusimple\cite{tusimple}, and LLAMAS\cite{behrendt2019unsupervised}. 
The experiment results show our method achieves state-of-the-art accuracy on all datasets.
The main contributions can be summarized as follows:
\begin{itemize}
    \item We demonstrate low-level and high-level features are complementary for lane detection, and we propose a novel network architecture~(CLRNet) to fully utilize low-level and high-level features for lane detection. 
	\item We propose ROIGather to further enhance the representation of lane features by gathering global context, which can also be plugged into other networks.
    \item We propose Line IoU~(LIoU) loss tailored for lane detection, regressing the lane as the whole unit and considerably improving the performance. 
    \item To better compare the localization accuracy of different detectors, we also adopt the new mF1 metrics. We demonstrate the proposed method greatly outperforms other state-of-the-art approaches on three lane detection benchmarks.
\end{itemize}
\section{Related Work}
\label{sec:related-work}
According to the representation of lane, current CNN-based lane detection can be divided into three categories: segmentation-based method, anchor-based method, and parameter-based method.

\subsection{Segmentation-based methods}
Modern algorithms typically adopt a pixel-wise prediction formulation, \emph{i.e.}, treat lane detection as a semantic segmentation task. SCNN~\cite{pan2018spatial} proposes a message-passing mechanism to address no visual evidence problem, which captures the strong spatial relationship for lanes. SCNN significantly improves the lane detection performance,  but the method is slow for real-time application. RESA~\cite{zheng2021resa} proposes a real-time feature aggregation module, enabling the network to gather the global feature and improve performance.
In CurveLane-NAS~\cite{xu2020curvelane}, they use neural architecture search~(NAS) to find a better network for capturing accurate information to benefit the detection of curve lanes. However, the NAS is extremely expensive computationally and costs huge GPU hours.
These segmentation-based methods are ineffective and time-consuming since 
they perform pixel-wise prediction on the whole image and don't consider lanes as a whole unit.

\subsection{Anchor-based methods}
Anchor-based methods in lane detection can be divided into two classes, \emph{e.g.}, line anchor-based methods and row anchor-based methods. 
Line anchor-based methods adopt predefined line anchors as references to regress accurate lanes.  Line-CNN~\cite{li2019line} is the pioneering work to use line anchors in lane detection.
LaneATT~\cite{tabelini2021keep} proposes a novel anchor-based attention mechanism that aggregates global information. It achieves state-of-the-art results and shows both high efficacy and efficiency. SGNet~\cite{su2021structure} introduces a novel vanish-point guided anchor generator and adds multiple structural guidance to improve performance. As for the row anchor-based method, it predicts the probable cell for each predefined row on images. UFLD~\cite{qin2020ultra} first proposes a row anchor-based lane detection method and adopts lightweight backbones to achieve high inference speed. Albeit simple and fast, its overall performance is not good. CondLaneNet~\cite{liu2021condlanenet} introduces a conditional lane detection strategy based on conditional convolution and row anchor-based formulation, \emph{i.e.}, it first locates start points of lane lines then performs row anchor-based lane detection. However, start points are hard to recognize in some complex scenarios, which results in relatively inferior performance.

\begin{figure*}[ht]
\centering
\includegraphics[width=0.95\linewidth]{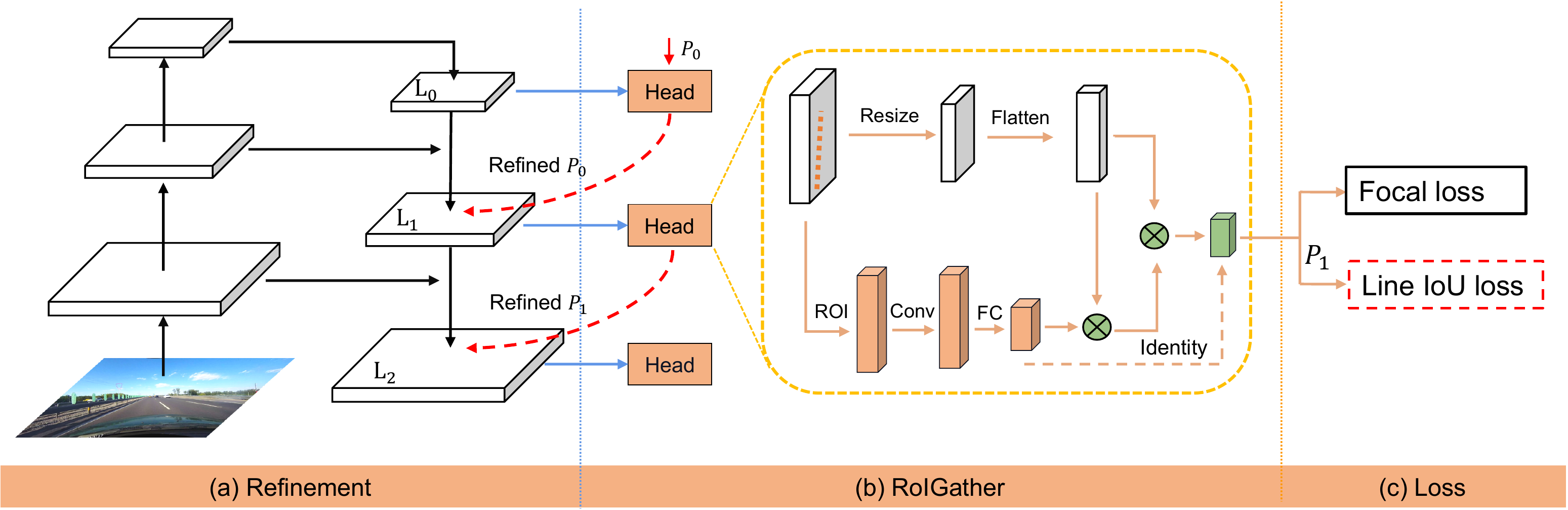}
\caption{Overview of the proposed \textbf{\methodname}. (a) The network generates feature maps from FPN\cite{lin2017feature} structure.  Subsequently, each lane prior will be refined from high-level features to low-level features. (b) Each head will exploit more contextual information for lane prior features. (c) Classification and regression of lane priors. The proposed Line IoU loss helps further improve the regression performance.
}
\label{fig:arch}
\end{figure*}

\subsection{Parameter-based methods}
Different from points regression, parameter-based methods model the lane curve with parameters and regress these parameters to detect lanes.
PolyLaneNet~\cite{tabelini2021polylanenet} adopts a polynomial regression problem and achieves high efficiency.  LSTR~\cite{liu2021end} takes road structures with camera pose into account to model the lane shape, then introduces the transformer to lane detection task to get the global feature. Parameter-based methods have fewer parameters to regress, but they are sensitive to the predicted parameters, \emph{e.g.}, the error prediction on high-order coefficient may cause shape change of lanes. Though parameter-based methods have fast inference speed, they still struggle to achieve higher performance.

\section{Approach}
\label{sec:approach}
\subsection{The Lane Representation}

\paragraph{Lane Prior.} Lanes are thin and long with strong shape priors, thus a predefined lane prior can help the network better localize lanes.
In common object detection, objects are represented by rectangular boxes. Nevertheless, the box is not appropriate for the representation
of the long line. Following \cite{li2019line} and \cite{tabelini2021keep}, we use equally-spaced 2D-points as lane representation. Specifically, lane is expressed
as a sequence of points, \emph{i.e.}, $P = \{(x_1, y_1), \cdots, (x_{N}, y_{N}) \}$. The $y$ coordinate of points is equally sampled through image vertically, \emph{i.e.},
$y_i = \frac{H}{N - 1} * i$, where $H$ is image height. Accordingly, the x coordinate is associated with the respective $y_i \in Y$. In our paper, we call this representation \textbf{Lane Prior}. Each lane prior will be predicted by the network and consists of four components: (1) foreground and background probabilities. (2) the length of lane prior. (3) the start point of the lane line and the angle between the x-axis of the lane prior~(termed as $x,y$, and $\theta$). (4) The $N$ offsets, \emph{i.e.}, the horizontal distance between the prediction and its ground truth.


\subsection{Cross Layer Refinement}\label{sec:clr}

\paragraph{Motivation.} 
In neural networks, deep high-level features strongly respond to entire objects with more semantic meanings, while the shallow low-level features are with more local contextual information. Allowing lane objects to access high-level features can help exploit more useful context information, \emph{e.g.}, to distinguish lane lines or landmarks. In the meantime, fine-detail features help detect lanes with high localization accuracy. In object detection~\cite{lin2017feature}, it builds the feature pyramid to leverage the pyramidal shape of a ConvNet's feature hierarchy and assigns different scales of objects to different pyramid levels. However, it is hard to directly assign a lane to only one level since high-level and low-level features are both critical for lanes. Inspired by Cascade RCNN~\cite{cai2018cascade}, we can assign lane objects to all levels and detect lanes sequentially. In particular, we can detect lanes with high-level features to localize lanes coarsely. Based on the detected lanes, we can refine them with more detailed features. 

\paragraph{Refinement structure.} 
Our goal is to leverage a ConvNet’s pyramidal feature hierarchy, which has semantics from low to high levels, and
build a feature pyramid with high-level semantics throughout. 
We take ResNet~\cite{he2016deep} as the backbone and use $\{L_0, L_1, L_2 \}$ to denote feature levels generated by FPN.
As shown in Fig.~\ref{fig:arch}, our cross layer refinement starts from the highest level $L_0$ and gradually approaches $L_2$.
We use $\{R_0, R_1, R_2\}$ to denote the corresponding refinements. 
Then we can build a sequence of refinements
\begin{align}
P_t = P_{t-1} \circ R_t(L_{t-1}, P_{t-1}), 
\end{align}
where $t= 1, \cdots, T$, T is the total number of refinements. Our method performs detection from highest level layer with high semantics. $P_t$ is the parameter of lane prior ~(start point coordinate $x, y$ and angle $\theta$), which is learnable inspired by~\cite{sparsercnn}. For the first layer $L_0$, the $P_0$ is uniformly distributed on image plane. The refinement $R_t$ takes the $P_t$ as input to get the ROI lane features and then performs two FC layers to get the refined parameter $P_t$.  
Progressively refining the lane prior and feature extraction is important for the success of cross layer refinement. Note that, our method is not limited to FPN structure, only using ResNet\cite{he2016deep} or adopting PAFPN~\cite{liu2018path} is also suitable.

\subsection{ROIGather}

\paragraph{Motivation.}

After we assign lane priors to each feature map, we can get features of lane priors with ROIAlign~\cite{he2017mask}. However, the contextual information of these features is still not sufficient. In some cases, the lane instance may be occupied or blurred with extreme lighting conditions. Thus there is no local visual evidence for the presence of lane. To determine whether a pixel belongs to a lane, we need to look at nearby features. Some recent studies~\cite{zhao2017pyramid, wang2018non} also indicate that the performance could be improved if making sufficient use of long-range dependencies. Thus, we can gather more useful contextual information to better learn lane feature. To this end, we add convolutions along the lane prior. 
In this way, each pixel in the lane prior can gather information of nearby pixels, and occupied parts can be reinforced from that information. 
Moreover, we build relations between features of lane priors and the whole feature map. Thus, it can exploit more contextual information to learn better feature representations.

\paragraph{ROIGather structure.}
The ROIGather module is light-weighted and easy to implement.
It takes feature map 
and lane priors as input, each lane prior has $N$ points. For each lane prior, we follow ROIAlign\cite{he2017mask} to get the ROI feature of lane prior ($\mathcal{X}_p \in \mathbb{R} ^{ C \times N_p }$). Unlike ROIAlign for bounding box, we uniform sample $N_p$ points from the lane prior and use bilinear interpolation to compute the exact values of input features at these locations. For ROI features of $L_1, L_2$, we concatenate the ROI features of previous layers to enhance feature representations. 
Convolutions are performed on the extracted ROI features to gather nearby features for each lane pixel.
To save memory, we use fully-connected to further extract the lane prior feature~($\mathcal{X}_p  \in \mathbb{R} ^{ C \times 1}$).  The feature map is resized to $\mathcal{X}_f \in \mathbb{R} ^ {C \times H \times W}$ and flattened to $\mathcal{X}_f \in \mathbb{R} ^ {C \times HW}$. Detail settings are in Sec.~\ref{sec:impl}.

To gather the global context for features of lane priors,  we first compute the attention\cite{wang2018non} matrix $\mathcal{W}$ between ROI lane prior feautre~($\mathcal{X}_p$) and the global feature map~($\mathcal{X}_f$), which is written as:
\begin{align}
\mathcal{W} = f(\frac{\mathcal{X}_p^T \mathcal{X}_f}{\sqrt{C}}),\label{attention}
\end{align}
where $f$ is a normalize function $softmax$. The aggregated feature is written as:
\begin{align}
\mathcal{G} = \mathcal{W} \mathcal{X}_f^T.
\end{align}

The output $\mathcal{G}$ reflects the bonus of $\mathcal{X}_f$ to $\mathcal{X}_p$ which is selected from all locations of $\mathcal{X}_f$. Finally, we add the output to the original input $X_{p}$.

\subsection{Line IoU loss}

\begin{figure}[t]
\centering
\includegraphics[width=0.99\linewidth]{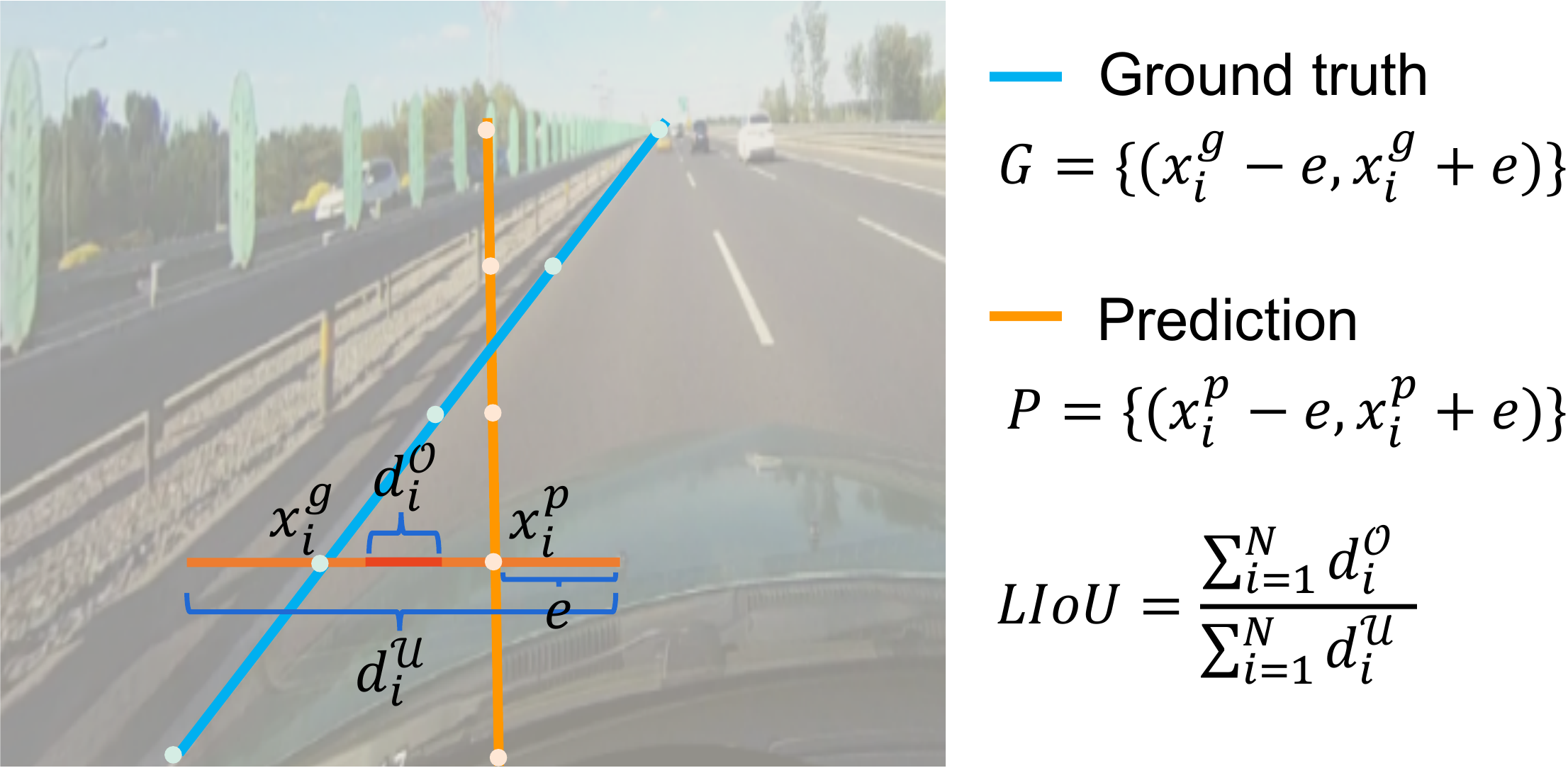}
\caption{
Illustration of Line IoU. Line IoU (interaction over union) can be calculated by integrating the IoU of the extended segment in terms of sampled $x_i$ position.
}
\label{fig:line_iou}
\end{figure}
\paragraph{Motivation.}
As discussed above, the lane prior consists of discrete points needed to be regressed with its ground truth. The commonly used distance loss like smooth-$l_1$ can be used to regress these points. However, this kind of loss takes points as separate variables, which is an oversimplified assumption~\cite{yu2016unitbox}, resulting in less accurate regression. In contrast to distance loss, Intersection over Union~(IoU) can take the lane prior as a whole unit to regress and it is tailored for evaluation metric\cite{yu2016unitbox,rezatofighi2019generalized, zheng2021scaloss}. In our work, we derive an easy and effective algorithm to compute the Line IoU~(LIoU) loss.

\paragraph{Formula.} We introduce Line IoU loss starting from the definition of the line segment IoU, which is the ratio of interaction over union between two line segments. For each point in the predicted lane as shown in Fig.~\ref{fig:line_iou}, we first extend it~($x_i^p$) with a radius $e$ into a line segment. Then IoU can be calculated between the extended line segment and its ground truth, which is written as:
\begin{equation}
\resizebox{0.9\linewidth}{!}{%
    $\begin{aligned}[b]
    IoU = \frac{ d_i^{\text{\tiny $\mathcal{O}$}}}{d_i^{\text{\tiny $\mathcal{U}$}}} = \frac{\mathrm{min}(x^p_i + e, x^g_i + e) - \mathrm{max}(x^p_i-e, x^g_i-e)}{\mathrm{max}(x^p_i + e, x^g_i + e) - \mathrm{min}(x^p_i-e, x^g_i-e)} ,
    \end{aligned}$
    }
\end{equation}

\noindent where $x^p_i-e, x^p_i + e$ are the extended points of $x^p_i$, $x^g_i - e, x^g_i + e$ are the corresponding ground truth points. Note that, $d_i^{\text{\tiny $\mathcal{O}$}}$ can be negative, which can make it feasible to optimize in case of non-overlapping line segments.

Then LIoU can be considered as the combination of infinite line points. To simplify the expression and make it easy to compute, we transform it into a discrete form,
\begin{align}
    LIoU = \frac{\sum_{i=1}^N d_i^{\text{\tiny $\mathcal{O}$}}}{\sum_{i=1}^N d_i^{\text{\tiny $\mathcal{U}$}}}.
\end{align}
Then, the LIoU loss is defined as
\begin{align}
    \mathcal{L}_{LIoU} = 1 - LIoU,
\end{align}
where $ -1 \leq LIoU \leq 1$, when two lines overlay perfectly, then $LIoU = 1$, $LIoU$ converges to -1 when two lines are far away. 

Our Line IoU loss exhibits two advantages: (1) It is simple and differentiable, which is very easy to implement parallel computations. (2) It predicts the lane as a whole unit, which helps improve the overall performance.

\subsection{Training and Infercence Details}


\paragraph{Positive samples selection.} 
During training, each ground truth lane is assigned with one or more predicted lanes dynamically as positive samples, which is inspired by~\cite{yolox}. In particular, we first sort the predicted lanes based on the assigning cost, which is defined as: 
\begin{equation}
\begin{aligned}
    \mathcal{C}_{assign} &= w_{sim} \mathcal{C}_{sim} + w_{cls} \mathcal{C}_{cls}, \\
    \mathcal{C}_{sim} &= (\mathcal{C}_{dis} \cdot \mathcal{C}_{xy} \cdot \mathcal{C}_{theta})^ 2 .
\end{aligned}
\end{equation}
Here $\mathcal{C}_{cls}$ is the focal cost\cite{focalloss} between predictions and labels. $\mathcal{C}_{sim}$ is the similarity cost between predicted lanes and ground truth. It consists of three parts, $\mathcal{C}_{dis}$ means the average pixel distance of all valid lane points, $\mathcal{C}_{xy}$ means the distance of start point coordinates, $\mathcal{C}_{theta}$ means the difference of the theta angle, they are all normalized to $[0, 1]$. $w_{cls}$ and $w_{sim}$ are weight coefficients of each defined component. Each ground truth lane is assigned with a dynamic number (top-k) of predicted lanes based on $\mathcal{C}_{assign}$.

\paragraph{Training Loss.} Training loss consists of classification loss and regression loss. The regression loss is only performed on the assigned samples. 
The overall loss function is defined as:
\begin{align}
    \mathcal{L}_{total} = w_{cls}\mathcal{L}_{cls} + w_{xytl}\mathcal{L}_{xytl} + w_{LIoU}\mathcal{L}_{LIoU}.
\end{align}
$\mathcal{L}_{cls}$ is the focal loss between predictions and labels, $\mathcal{L}_{xytl}$ is the smooth-$l_1$ loss for the start point coordinate, theta angle and lane length regression, $\mathcal{L}_{LIoU}$ is the Line IoU loss between the predicted lane and ground truth. Optionally, we can add an auxiliary segmentation loss following\cite{qin2020ultra}. It is only used in the training period and has no cost in inference.


\paragraph{Inference.} We set a threshold with a classification score to filter
the background lanes (low score lane priors), and we use
nms to remove high-overlapped lanes following \cite{tabelini2021keep}.
Our method can also be nms-free if we use
the one-to-one assignment, i.e., set the top-k = 1.

\section{Experiment}
\label{sec:exp}

\subsection{Datasets}
We conduct experiments on two widely used lane detection benchmark datasets: \textbf{CULane}~\cite{pan2018spatial} and \textbf{Tusimple}~\cite{tusimple} and
one recently released benchmark~(\textbf{LLAMAS}~\cite{behrendt2019unsupervised}).

\textbf{CULane}~\cite{pan2018spatial} is a large scale challenging dataset for lane detection. It contains nine challenging categories, such as crowded, night, cross, etc. The CULane dataset consists of 100,000 images for train, validation, and test sets. All the images have $1640 \times 590$ pixels. 

\textbf{LLAMAS}\cite{behrendt2019unsupervised} is also a large scale lane detection dataset with over 100k images. The lane markers in LLAMAS are automatically annotated with highly-accurate maps. Since the label of test set is not public, we upload the detection result to the website of LLAMAS benchmark for testing.

\textbf{Tusimple}~\cite{tusimple} lane detection benchmark is one of the most widely used datasets in lane detection. It contains only highway scenes with 3268 images for training, 358 for validation, and 2782 for testing. All have $1280 \times 720$ pixels.

\subsection{Implementation details} 
We adopt the ResNet~\cite{he2016deep} and DLA\cite{yu2018deep} as our pre-trained backbones. All input images are resized to $320 \times 800$. For data augmentation, similar to\cite{liu2021condlanenet, qu2021focus}, we use random affine transformation~(translation, rotation, and scaling), random horizontal flips. In the optimizing process, we use AdamW~\cite{loshchilov2018decoupled} optimizer with an initial learning rate of 1e-3 and cosine decay learning rate strategy~\cite{loshchilov2016sgdr} with power set to 0.9. We train 15 epochs, 70 epochs, 20 epochs for CULane, Tusimple, and LLAMAS, respectively.
Our network is implemented based on Pytorch with 1GPU to run all the experiments.
We set the number points of lane prior $N=72$, and the sampled number $N_p=36$. The resized $H, W$ in ROIGather are $10, 25$, respectively, channel $C=64$. The extended radius $e$ in $LIoU$ is 15. The coefficients of assigning cost are set as $w_{cls} = 1$ and $w_{sim} = 3$.
\label{sec:impl}

\subsection{Evaluation Metric}
\begin{table*}

    \begin{center}

        \resizebox{\textwidth}{!}{%

            \begin{tabular}{@{}lrrrrrrrrrrrrrrr@{}}

                \toprule

                \multicolumn{1}{c}{\textbf{Method}} & \multicolumn{1}{c}{\textbf{Backbone}} & \multicolumn{1}{c}{\textbf{mF1}} &
                \multicolumn{1}{c}{\textbf{F1@50}} &
                \multicolumn{1}{c}{\textbf{F1@75}} &
                \multicolumn{1}{c}{\textbf{FPS}} &
                \multicolumn{1}{c}{\textbf{GFlops} } &
                \multicolumn{1}{c}{\textbf{Normal}} & \multicolumn{1}{c}{\textbf{Crowded}} & \multicolumn{1}{c}{\textbf{Dazzle}} & \multicolumn{1}{c}{\textbf{Shadow}} & \multicolumn{1}{c}{\textbf{No line}} & \multicolumn{1}{c}{\textbf{Arrow}} & \multicolumn{1}{c}{\textbf{Curve}} & \multicolumn{1}{c}{\textbf{Cross}} & \multicolumn{1}{c}{\textbf{Night}} \\ \midrule

                SCNN~\cite{pan2018spatial} & VGG16 & 38.84 & 71.60 & 39.84 & 7.5 & 328.4 & 90.60 & 69.70 & 58.50 & 66.90 & 43.40 & 84.10 & 64.40 & 1990 & 66.10 \\
                
                RESA~\cite{zheng2021resa} & ResNet34 & - & 74.50 & - & 45.5 & 41.0 & 91.90 & 72.40 & 66.50 & 72.00 & 46.30 & 88.10 & 68.60 & 1896 & 69.80  \\

                RESA~\cite{zheng2021resa} & ResNet50 & 47.86 & 75.30 & 53.39 & 35.7 & 43.0 & 92.10 & 73.10 & 69.20 & 72.80 & 47.70 & 88.30 & 70.30 & 1503 & 69.90  \\

                FastDraw~\cite{philion2019fastdraw} & ResNet50 & - & - & - & 90.3 & - & 85.90 & 63.60 & 57.00 & 69.90 & 40.60 & 79.40 & 65.20 & 7013 & 57.80  \\

                E2E~\cite{yoo2020end} & ERFNet & - & 74.00 & - & - & - & 91.00 & 73.10 & 64.50 & 74.10 & 46.60 & 85.80 & 71.90 & 2022 & 67.90 \\
                
                UFLD~\cite{qin2020ultra} & ResNet18 & 38.94 & 68.40 & 40.01 & \textbf{282} & \textbf{8.4} & 87.70 & 66.00 & 58.40 & 62.80 & 40.20 & 81.00 & 57.90 & 1743 & 62.10 \\

                UFLD~\cite{qin2020ultra} & ResNet34 & - & 72.30 & - & 170 & 16.9 & 90.70 & 70.20 & 59.50 & 69.30 & 44.40 & 85.70 & 69.50 & 2037 & 66.70 \\
                
                PINet~\cite{ko2021key} & Hourglass & 46.81 & 74.40 & 51.33 & 25 & - & 90.30 & 72.30 & 66.30 & 68.40 & 49.80 & 83.70 & 65.20 & 1427 & 67.70 \\

                LaneATT~\cite{tabelini2021keep} & ResNet18 & 47.35 & 75.13 & 51.29 & 153 & 9.3 & 91.17 & 72.71 & 65.82 & 68.03 & 49.13 & 87.82 & 63.75 & \textbf{1020} & 68.58 \\

                LaneATT~\cite{tabelini2021keep} & ResNet34 & 49.57 & 76.68 & 54.34 & 129 & 18.0 & 92.14 & 75.03 & 66.47 & 78.15 & 49.39 & 88.38 & 67.72 & 1330 & 70.72 \\

                LaneATT~\cite{tabelini2021keep} & ResNet122 & 51.48 & 77.02 & 57.50 & 20 & 70.5 & 91.74 & 76.16 & 69.47 & 76.31 & 50.46 & 86.29 & 64.05 & 1264 & 70.81 \\
                
                LaneAF~\cite{abualsaud2021laneaf} & ERFNet & 48.60 & 75.63 & 54.53 & 24 & 22.2 & 91.10 & 73.32 & 69.71 & 75.81 & 50.62 & 86.86 & 65.02 & 1844 & 70.90 \\
                
                LaneAF~\cite{abualsaud2021laneaf} & DLA34 & 50.42 & 77.41 & 56.79 & 20 & 23.6 & 91.80 & 75.61 & 71.78 & 79.12 & 51.38 & 86.88 & 72.70 & 1360 & 73.03 \\
                
                SGNet~\cite{su2021structure} & ResNet18 & - & 76.12 & - & 117 & - & 91.42 & 74.05 & 66.89 & 72.17 & 50.16 & 87.13 & 67.02 & 1164 & 70.67 \\
                
                SGNet~\cite{su2021structure} & ResNet34 & - & 77.27 & - & 92 & - & 92.07 & 75.41 & 67.75 & 74.31 & 50.90 & 87.97 & 69.65 & 1373 & 72.69 \\
                
                FOLOLane~\cite{qu2021focus} & ERFNet & - & 78.80 & - & 40 & - & 92.70 & 77.80 & 75.20 & 79.30 & 52.10 & 89.00 & 69.40 & 1569 & 74.50 \\
                
                CondLane~\cite{liu2021condlanenet} & ResNet18 & 51.84 & 78.14 & 57.42 & 173 & 10.2 & 92.87 & 75.79 & 70.72 & 80.01 & 52.39 & 89.37 & 72.40 & 1364 & 73.23 \\

                CondLane\cite{liu2021condlanenet} & ResNet34 & 53.11 & 78.74 & 59.39 & 128 & 19.6 & 93.38 & 77.14 & 71.17 & 79.93 & 51.85 & 89.89 & 73.88 & 1387 & 73.92 \\

                CondLane\cite{liu2021condlanenet} & ResNet101 & 54.83 & 79.48 & 61.23 & 47 & 44.8 & 93.47 & 77.44 & 70.93 & 80.91 & 54.13 & 90.16 & 75.21 & 1201 & 74.80 \\

                \midrule

                \textbf{\methodname~(ours)} & ResNet18 &  55.23 & 79.58 & 62.21  & 119/206* & 11.9 & 93.30 & 78.33 &  73.71 & 79.66 &  53.14 &  90.25 &  71.56 & 1321 & 75.11 \\

                \textbf{\methodname~(ours)} & ResNet34 & 55.14 &  79.73 &  62.11 & 103/156* & 21.5 &  93.49 & 78.06 & 74.57 & 79.92 & 54.01 & 90.59 & 72.77 &1216 & 75.02\\

                \textbf{\methodname~(ours)} & ResNet101 & 55.55 & 80.13 & \textbf{62.96} & 46/74* & 42.9 & \textbf{93.85} & 78.78 & 72.49 & 82.33 & 54.50 & 89.79 & \textbf{75.57} & 1262 & \textbf{75.51} \\
                
                \textbf{\methodname~(ours)} & DLA34 & \textbf{55.64} & \textbf{80.47} & 62.78 & 94/151* & 18.5 & 93.73 & \textbf{79.59} & \textbf{75.30} & \textbf{82.51} & \textbf{54.58} & \textbf{90.62} & 74.13 &1155 & 75.37 \\
                
                \bottomrule
            \end{tabular}

        } %

    \end{center}
    \caption{
    State-of-the-art results on CULane. For a fairer comparison, we remeasure the FPS of the source code available detectors using one NVIDIA 1080Ti GPU on the same machine, * means FPS on TensorRT. In addition, we also evaluation these detectors to report the mF1, F1@50, F1@75. For ``Cross'' category , only false positives are shown. The reported metric of these categories is based on F1@50.
    }

    \label{tab:culane_main}

\end{table*}

We adopt the F1-measure as evaluation metric for CULane~\cite{pan2018spatial} and LLAMAS~\cite{behrendt2019unsupervised}. Intersection-over-union~(IoU) is calculated between predictions and ground truth. Predicted lanes whose IoU are larger than a threshold~(0.5) are considered as true positives~(TP). The $F_1$ is defined as:
$$
F_1 = \frac{2\times Precision \times Recall}{Precision + Recall}.
$$
Following COCO~\cite{lin2014microsoft} detection metric, we also report a new metric mF1 to better compare the localization performance of algorithms. It is defined as 
$$
\mathrm{mF1} = (\mathrm{F1@50 + F1@55 + \cdots + F1@95}) / 10,
$$
where F1@50, F1@55, $\cdots$, F1@95 are F1 metrics when IoU thresholds are $0.5, 0.55, \cdots, 0.95$ respectively. This is a break from the tradition which will reward detectors with better localization results.

For Tusimple\cite{tusimple} dataset, the evaluation formula is
$$
Accuracy = \frac{\sum_{clip}C_{clip}}{\sum_{clip}S_{clip}},
$$
where $C_{clip}, S_{clip}$ are the number of correct points and the number of ground truth points of a image respectively. A predicted lane is a correct one if more than 85\% predicted lane points are
within 20 pixels the ground truth. Tusimple dataset also reports the rate of false positive~(FP) and false negative~(FN), where $FP = \frac{F_{pred}}{N_{pred}}, FN = \frac{M_{pred}}{N_{gt}}$.

\subsection{Comparison with the state-of-the-art results}
\paragraph{Performance on CULane.}
We show the results of our method on the CULane lane detection benchmark dataset and compare them with other popular lane detection methods. As illustrated in Table~\ref{tab:culane_main}, our proposed method achieves a new state-of-the-art on CULane with an 80.47 F1@50 measure. The ResNet18 version of our method achieves 79.58 F1@50, which is even higher than CondLaneNet~(ResNet101) while getting 1.4 points higher than CondLaneNet~(ResNet18). In particular, we surpass CondLaneNet~(ResNet18) by 3.4\% mF1, which indicates our method better regresses lanes with high localization accuracy. Comparing line anchor-based method LaneATT, our ResNet18 version surpasses 7.88 \% mF1 and 4.45 \% F1@50, respectively. In the meantime, CLRNet can achieve 206 FPS in one NVIDIA 1080Ti GPU with TensorRT, which is efficient for real-time lane detection.

\begin{figure*}[h]
  \centering
  \includegraphics[width=0.95\linewidth]{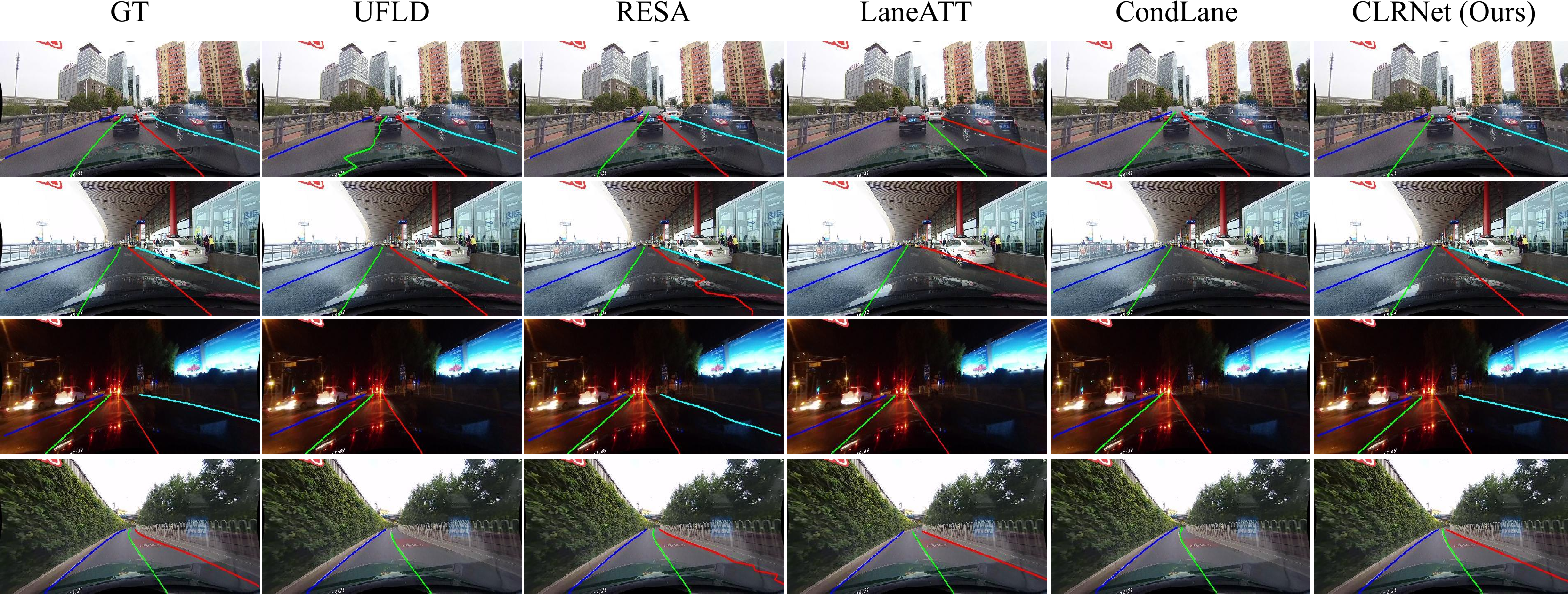}
  \caption{Visualization results of UFLD, RESA, LaneATT, CondLane and our method on CULane testing set.}
  \label{fig:vis-comparison}
\end{figure*}

We show the qualitative results on the CULane dataset in Fig.~\ref{fig:vis-comparison}. Segmentation-based methods like RESA don't predict the lane as a whole unit, which can not preserve the smoothness of lanes. CondLaneNet only predicts one start point of the lane as the proposal, it is easy to miss some lane instances. Our method can predict continuous and smooth lanes in these challenging scenarios, which demonstrates our method can definitely gather global context and has a strong ability to detect accurate lanes.


\paragraph{Performance on LLAMAS.}The result on the LLAMAS dataset is shown in Table~\ref{tab:llamas_main}. Our method outperforms PolyLaneNet~\cite{tabelini2021polylanenet} and LaneATT~\cite{tabelini2021keep} by 7.7 F1@50 and 2.4 F1@50 respectively on the test set, which is significant improvement. Although LaneAF\cite{abualsaud2021laneaf} achieves 96.90 F1@50 in the valid dataset, its inference speed is slow~(near 20FPS), which makes it hard for deployment. Moreover, our method achieves near 2 points mF1 higher than LaneAF, which demonstrates our method is more accurate in localization.

\begin{table}[t]
\resizebox{0.5\textwidth}{!}{%

\begin{tabular}{@{}c c c c c c c c c c c@{}} \toprule
\multirow{2}{*}{\textbf{Method}} & \multirow{2}{*}{\textbf{Backbone}} &\multicolumn{3}{c}{\textbf{valid}} & \textbf{test} \\ \cmidrule(lr){3-5} \cmidrule(lr){6-6} 
& & \textbf{mF1} & \textbf{F1@50} & \textbf{F1@75} & \textbf{F1@50} \\\midrule
PolyLaneNet\cite{tabelini2021polylanenet} & EfficientnetB0  & 48.82 & 90.2 &  45.40 & 88.40\\
LaneATT\cite{tabelini2021keep} & ResNet18  &  69.22 & 94.64 & 82.36 & 93.46 \\
LaneATT\cite{tabelini2021keep} & ResNet34 & 69.63 & 94.96 & 82.79 & 93.74 \\
LaneATT\cite{tabelini2021keep} & ResNet122 &  70.8 & 95.17 & 84.01& 93.54 \\
LaneAF\cite{abualsaud2021laneaf} & DLA34 & 69.31 & 96.90 & 84.71 & 96.07\\ \midrule
\textbf{\methodname~(ours)} & ResNet18 & \textbf{71.61} & 96.96 & \textbf{85.59} & 96.00 \\
 \textbf{\methodname~(ours)} & DLA34 & 71.21 & \textbf{97.16} & 85.33 & \textbf{96.12} \\\bottomrule
\end{tabular}
}
\caption{State-of-the-art results on LLAMAS. Additionally, we rerun the evaluation for these methods with source code and trained models to get the mF1, F1@50, F1@75.}
\label{tab:llamas_main}
\end{table}

\paragraph{Performance on Tusimple.} Table~\ref{tab:tusimple_main} shows the  performance comparison with state-of-the-art approaches. The performance difference between different methods on this dataset is very small,  which shows the result in this dataset seems to be saturated (high value) already. Our method achieves a new start-of-the-art in terms of F1 score and surpasses the previous state-of-the-art with a 0.6\% F1 score. This significant improvement manifests the effectiveness of our method.

\begin{figure*}[t]
	\centering
    \includegraphics[width=0.95\linewidth]{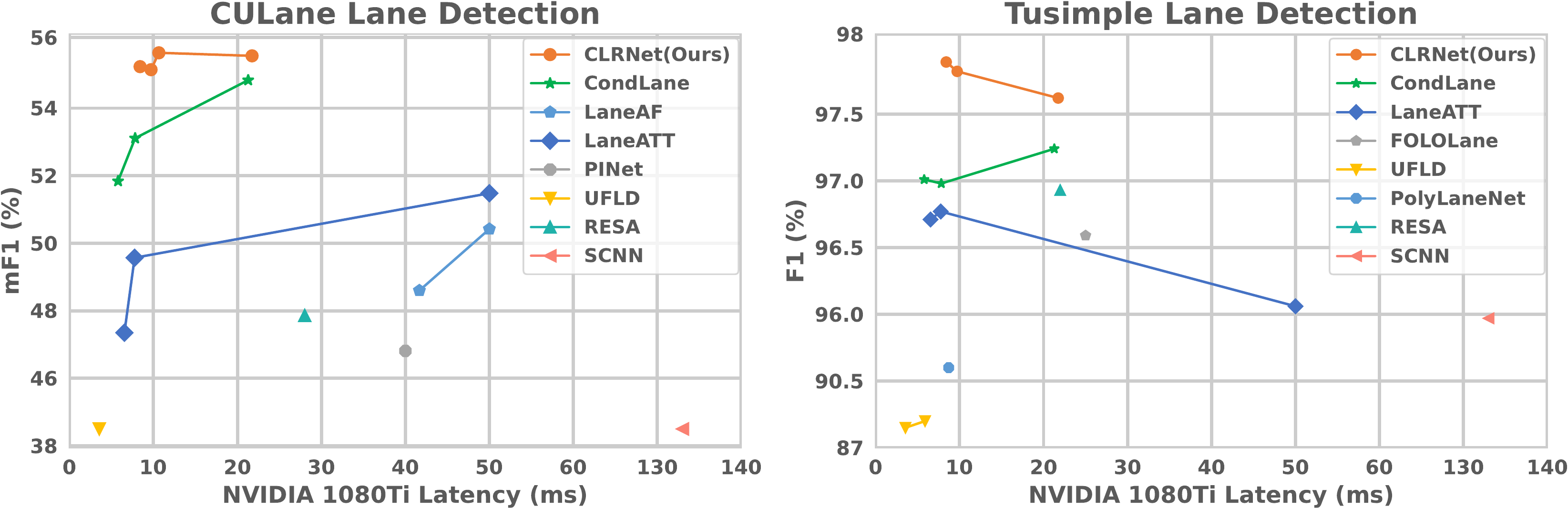}
	\vspace{-5pt}
	\caption{Latency vs. F1-score of state-of-the-art methods on CULane and Tusimple benchmarks.}
	\label{figs:latency-mf1}
	\vspace{-7pt}
\end{figure*}

\begin{table}
    \begin{center}
    \resizebox{0.5\textwidth}{!}{%
            \begin{tabular}{@{}lrrrrrrr@{}}
                \toprule
                \textbf{Method} & \textbf{Backbone}        & \textbf{F1 (\%)} & \textbf{Acc (\%)}  & \textbf{FP (\%)} & \textbf{FN (\%)} \\ \midrule
                
                SCNN~\cite{pan2018spatial} & VGG16 & 95.97 & 96.53  & 6.17 & \textbf{1.80} \\
                
                RESA~\cite{zheng2021resa} & ResNet34 & 96.93 & 96.82 & 3.63 & 2.48  \\
                
                PolyLaneNet~\cite{tabelini2021polylanenet}  & EfficientNetB0 & 90.62 & 93.36 & 9.42 & 9.33 \\
                
                E2E~\cite{yoo2020end}  & ERFNet  & 96.25  & 96.02  & 3.21 & 4.28 \\
                
                UFLD\cite{qin2020ultra} & ResNet18 & 87.87 & 95.82 & 19.05 & 3.92 \\
                
                UFLD\cite{qin2020ultra}  &  ResNet34 & 88.02 & 95.86  & 18.91 & 3.75 \\
                
                LaneATT\cite{tabelini2021keep} & ResNet18 & 96.71 & 95.57 & 3.56 & 3.01 \\
                
                LaneATT\cite{tabelini2021keep} & ResNet34 & 96.77 & 95.63 & 3.53 & 2.92 \\
                
                LaneATT\cite{tabelini2021keep} & ResNet122 & 96.06 & 96.10 & 5.64 & 2.17 \\
                
                FOLOLane\cite{qu2021focus} & ERFNet & 96.59 & \textbf{96.92} & 4.47 & 2.28 \\
                
                CondLaneNet\cite{liu2021condlanenet} & ResNet18 & 97.01 & 95.48 & 2.18 & 3.80 \\
                
                CondLaneNet\cite{liu2021condlanenet} & ResNet34 & 96.98 & 95.37  & 2.20 & 3.82 \\
                
                CondLaneNet\cite{liu2021condlanenet} & ResNet101 & 97.24 & 96.54 & \textbf{2.01} & 3.50 \\
                \midrule

                \textbf{\methodname~(ours)} & ResNet18                                & \textbf{97.89}                                 & 96.84                                 & 2.28                                 & 1.92  \\

                \textbf{\methodname~(ours)}  &     ResNet34                & 97.82                                 &  96.87                                 & 2.27                                 &  2.08    \\

                \textbf{\methodname~(ours)} & ResNet101                               & 97.62                                 &  96.83                                 & 2.37                                 & 2.38    \\ 
                \bottomrule
            \end{tabular}
    }

    \end{center}
    \caption{State-of-the-art results on TuSimple. Additionally, F1 was computed using the official source code.}

    \label{tab:tusimple_main}

\end{table}

\begin{table}[htb]
	\centering
	
	\vspace{0.1cm}
	\addtolength{\tabcolsep}{0pt}
	\resizebox{0.48\textwidth}{!}{%
	\begin{tabular}{*{12}{c}}
		\toprule
		\textbf{LIoU} & \textbf{Refinement} & \textbf{ROIGather} & \textbf{mF1}   & \textbf{F1@50}  & \textbf{F1@75} & \textbf{F1@90} & \\
		\midrule
		                      &                          &                  &  51.90 & 78.37 & 58.32 & 14.43    \\
		\checkmark            &                          &                  &  52.80 & 78.27 &  59.50  & 16.54  \\
		\checkmark            & \checkmark               &                  &  54.74 & 78.91  & 61.77 & 20.09    \\
		\checkmark            & \checkmark               & \checkmark       & \textbf{55.23} & \textbf{79.58} & \textbf{62.21} & \textbf{20.64}  \\
		\bottomrule
	\end{tabular}
	}
	\caption{Effects of each component in our method. Results are reported on CULane.}
	\label{tab:overall-ablation}
	
\end{table}

\subsection{Ablation study}
To validate the effectiveness of different components of the proposed method, we conducted several experiments on the CULane dataset to show the performance.

\paragraph{Overall Ablation Study.} To analyze the importance of each proposed method, we report the overall ablation studies in Table \ref{tab:overall-ablation}. We gradually add LIoU loss, Cross Layer Refinement, and ROIGather on the ResNet18 baseline.  LIoU loss improves the mF1 from 51.90 to 52.80. This result validates that the localization accuracy is much improved. Moreover, the refinement further improves the mF1 to 54.74. Results in mF1, F1@50, F1@70, and F1@90 are consistently improved, which validates that leveraging high-level and low-level semantic features to detect lanes is useful and yields consistent improvements. ROIGather further improves mF1 by 0.5\%, which validates rich global context can enhance the representation of lane features.
\begin{figure}[t]
\centering
\includegraphics[width=0.95\linewidth]{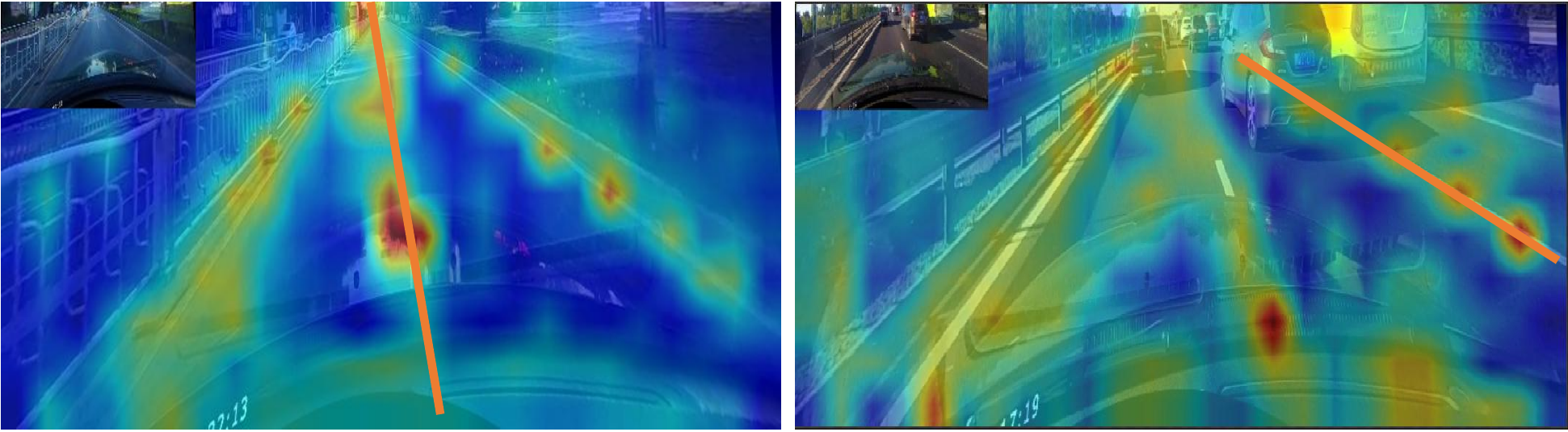}
\caption{Illustration of attention weight in ROIGather. It shows attention~(Eq.~\ref{attention}) between the ROI feature of the lane prior and the whole feature map. The {\color[rgb]{1,0.5,0} orange} line is the correspond lane prior. The {\color{red} red} regions corresponds to high score in attention weight. 
}
\label{figs:roi_gather}
\end{figure}

\paragraph{Analysis for ROIGather.} 
To further demonstrate how ROIGather works in the network, we visualize the attention map ~(Eq.~\ref{attention}) in Fig.~\ref{figs:roi_gather}. 
It shows attention weight between the ROI feature of the lane prior (orange line) and the whole feature map. The brighter the color is, the larger the weight value
is. Notably, the proposed ROIGather can (i) effectively gather global context with rich semantic information, (ii) capture the feature of foreground lanes even under occlusion. More quantitative results are in Appendix.



\begin{table}[t]
	\centering
	\resizebox{\linewidth}{!}{
	\begin{tabular}{*{12}{c}}
		\toprule
		\textbf{Settings}  & \textbf{mF1} & \textbf{F1@50} & \textbf{F1@75} & \textbf{F1@90}  \\
		\midrule
		$R_0$ & 52.80 & 78.27 &  59.50  & 16.54 \\
		$R_1$ & 52.26 & 78.05 & 59.27 & 16.71 \\
		$R_2$ & 52.50 & 77.82  & 59.05 & 16.82 \\
		\midrule
		$R_0$ $\rightarrow$ $R_0$ & 53.02 & 78.63 & 59.74 & 16.88 \\
		$R_0$ $\rightarrow$ $R_0$ $\rightarrow$ $R_0$ & 53.21 & 78.61 & 60.11 & 16.96 \\
		$R_2$ $\rightarrow$ $R_1$ $\rightarrow$ $R_0$ & 53.58 & 78.73 &  60.44  & 17.58 \\
		ADD & 53.77 & 78.52 & 60.45 & 19.01 \\
		\midrule
		$R_0$ $\rightarrow$ $R_1$ $\rightarrow$ $R_2$ & \textbf{54.74} & \textbf{78.91} & \textbf{61.77} & \textbf{20.09} \\
		\bottomrule
	\end{tabular}
	}
	\caption{Ablation studies of on different refinement methods. $R_i$ is the refinement method discussed in \cref{sec:clr}. \textit{ADD} means add all features with refinement iteration=3 for a fairer comparison.}
	\label{tab:refinement}
	
\end{table}

\paragraph{Ablation study on Cross Layer Refinement.}
Ablation studies of Cross Layer Refinement are shown in Table~\ref{tab:refinement}.
We first implement the detector with only one layer to perform refinement. As we can see from the result~(setting $R_0, R_1, R_2$), these three refinements get similar results. $R_2$ gets relatively high F1@90 while the F1@50 is relatively low, indicating low-level features help regress lanes accurately. However, it may cause false detection due to losing high semantic information. 
We select the better result $R_0$ and gradually add more refinements. As $R_0 \rightarrow R_0$ shows, it gets slightly improvement.
Other fusion feature methods like adding all features still cannot give an improvement.
Adopting the refinement from $R_0$ to $R_2$ is much better than others, which validates our cross layer refinement can utilize high-level and low-level features better. 




\paragraph{Ablation Study on Line-IoU Loss.}

Ablation studies of Line IoU loss are shown in Table~\ref{tab:line_iou}. We first turn loss weight to select the best regression weight of smooth-$l_1$.
We observe that the regression loss of smooth-$l_1$ is much larger than classification loss when regression weight is 1.5. Results show decreasing weight to 0.5 is relatively better. In contrast, LIoU loss is more stable and improves the performance by near 1 point mF1. To be more specific, the improvements mostly come from high overlapping metrics, like F1@80 and F1@90.
These experimental results validate that our Line IoU loss can achieve better performance and make the model better converged. We show the proposed Line IoU loss can also improve the performance of LaneATT\cite{tabelini2021keep}, details can be found in Appendix.


\begin{table}[t]
	\centering
	\resizebox{\linewidth}{!}{
	\begin{tabular}{*{12}{c}}
		\toprule
		\textbf{Loss} & \textbf{Weight} & \textbf{mF1} & \textbf{F1@50} & \textbf{F1@60} & \textbf{F1@70} & \textbf{F1@80} & \textbf{F1@90}  \\
		\midrule
		\multirow{3}{*}{smooth-$l_1$} 
		  & 0.1 & 54.15 & 79.23 & 75.00 & 67.35 & 51.47 & 18.98  \\
		  & 0.5 & 54.22 & 79.05 & 74.94 & 67.21 & 52.03 & 19.04 \\
		  & 1.0 & 53.48 & 78.33 & 74.09 & 66.63 & 50.85 & 18.52 \\
		  & 1.5 & 52.68 & 78.23 & 73.57 & 65.43 & 49.38 & 17.58 \\
		 \midrule
		 \multirow{3}{*}{$\mathcal{L}_{LIoU}$} & 2 & 55.23 & \textbf{79.58} & \textbf{75.53} & 68.26 & 53.62 & 20.64\\
		    & 4 & 55.22 & 79.31 & 75.43 & \textbf{68.54} & 53.78 & 20.38 \\
		    & 6 & \textbf{55.31} & 79.35 & 75.45 & 68.48 & \textbf{53.82} & \textbf{20.83}\\
		\bottomrule
	\end{tabular}
	}
	\caption{Ablation studies of Line IoU loss on CULane.}
	\label{tab:line_iou}
\end{table}

\section{Conclusion}
In this paper, we present Cross Layer Refinement Network~(CLRNet) for lane detection. CLRNet can exploit high-level features to predict lanes while leveraging local-detailed features to improve localization accuracy. To solve the no visual evidence for the presence of lane, we propose ROIGather to enhance the representation of lane features by building relations with all pixels. To regress lane as a whole unit, we propose Line IoU loss tailored for lane detection, which considerably improves the performance compared with standard loss, \emph{i.e.}, smooth-$l_1$ loss. Our method is evaluated on three lane detection benchmark datasets, \emph{i.e., }CULane, LLamas, and Tusimple. Experiments show our proposed method outperforms current state-of-the-art lane detection methods.

\section{Acknowledgement}
This work was supported in part by The National Key Research and Development Program of China (Grant Nos: 2018AAA0101400), in part by The National Nature Science Foundation of China (Grant Nos: 62036009, U1909203, 61936006, 62133013), in part by Innovation Capability Support Program of Shaanxi (Program No. 2021TD-05).

\label{sec:conclusion}


{\small
\bibliographystyle{ieee_fullname}
\bibliography{egbib}

\begin{thebibliography}{10}\itemsep=-1pt

\bibitem{abualsaud2021laneaf}
Hala Abualsaud, Sean Liu, David Lu, Kenny Situ, Akshay Rangesh, and Mohan~M
  Trivedi.
\newblock Laneaf: Robust multi-lane detection with affinity fields.
\newblock {\em arXiv preprint arXiv:2103.12040}, 2021.

\bibitem{behrendt2019unsupervised}
Karsten Behrendt and Ryan Soussan.
\newblock Unsupervised labeled lane marker dataset generation using maps.
\newblock In {\em International Conference on Computer Vision (ICCV)},
  volume~1, page~7, 2019.

\bibitem{cai2018cascade}
Zhaowei Cai and Nuno Vasconcelos.
\newblock Cascade r-cnn: Delving into high quality object detection.
\newblock In {\em Proceedings of the IEEE conference on computer vision and
  pattern recognition}, pages 6154--6162, 2018.

\bibitem{yolox}
Zheng Ge, Songtao Liu, Feng Wang, Zeming Li, and Jian Sun.
\newblock Yolox: Exceeding yolo series in 2021.
\newblock {\em arXiv preprint arXiv:2107.08430}, 2021.

\bibitem{he2017mask}
Kaiming He, Georgia Gkioxari, Piotr Doll{\'a}r, and Ross Girshick.
\newblock Mask r-cnn.
\newblock In {\em Proceedings of the IEEE international conference on computer
  vision}, pages 2961--2969, 2017.

\bibitem{he2016deep}
Kaiming He, Xiangyu Zhang, Shaoqing Ren, and Jian Sun.
\newblock Deep residual learning for image recognition.
\newblock In {\em Proceedings of the IEEE conference on computer vision and
  pattern recognition}, pages 770--778, 2016.

\bibitem{ko2021key}
Yeongmin Ko, Younkwan Lee, Shoaib Azam, Farzeen Munir, Moongu Jeon, and Witold
  Pedrycz.
\newblock Key points estimation and point instance segmentation approach for
  lane detection.
\newblock {\em IEEE Transactions on Intelligent Transportation Systems}, 2021.

\bibitem{li2019line}
Xiang Li, Jun Li, Xiaolin Hu, and Jian Yang.
\newblock Line-cnn: End-to-end traffic line detection with line proposal unit.
\newblock {\em IEEE Transactions on Intelligent Transportation Systems},
  21(1):248--258, 2019.

\bibitem{lin2017feature}
Tsung-Yi Lin, Piotr Doll{\'a}r, Ross Girshick, Kaiming He, Bharath Hariharan,
  and Serge Belongie.
\newblock Feature pyramid networks for object detection.
\newblock In {\em Proceedings of the IEEE conference on computer vision and
  pattern recognition}, pages 2117--2125, 2017.

\bibitem{focalloss}
Tsung-Yi Lin, Priya Goyal, Ross Girshick, Kaiming He, and Piotr Doll{\'a}r.
\newblock Focal loss for dense object detection.
\newblock In {\em Proceedings of the IEEE international conference on computer
  vision}, pages 2980--2988, 2017.

\bibitem{lin2014microsoft}
Tsung-Yi Lin, Michael Maire, Serge Belongie, James Hays, Pietro Perona, Deva
  Ramanan, Piotr Doll{\'a}r, and C~Lawrence Zitnick.
\newblock Microsoft coco: Common objects in context.
\newblock In {\em European conference on computer vision}, pages 740--755.
  Springer, 2014.

\bibitem{liu2021condlanenet}
Lizhe Liu, Xiaohao Chen, Siyu Zhu, and Ping Tan.
\newblock Condlanenet: a top-to-down lane detection framework based on
  conditional convolution.
\newblock {\em arXiv preprint arXiv:2105.05003}, 2021.

\bibitem{liu2021end}
Ruijin Liu, Zejian Yuan, Tie Liu, and Zhiliang Xiong.
\newblock End-to-end lane shape prediction with transformers.
\newblock In {\em Proceedings of the IEEE/CVF Winter Conference on Applications
  of Computer Vision}, pages 3694--3702, 2021.

\bibitem{liu2018path}
Shu Liu, Lu Qi, Haifang Qin, Jianping Shi, and Jiaya Jia.
\newblock Path aggregation network for instance segmentation.
\newblock In {\em Proceedings of the IEEE conference on computer vision and
  pattern recognition}, pages 8759--8768, 2018.

\bibitem{loshchilov2016sgdr}
Ilya Loshchilov and Frank Hutter.
\newblock Sgdr: Stochastic gradient descent with warm restarts.
\newblock {\em arXiv preprint arXiv:1608.03983}, 2016.

\bibitem{loshchilov2018decoupled}
Ilya Loshchilov and Frank Hutter.
\newblock Decoupled weight decay regularization.
\newblock In {\em International Conference on Learning Representations}, 2018.

\bibitem{pan2018spatial}
Xingang Pan, Jianping Shi, Ping Luo, Xiaogang Wang, and Xiaoou Tang.
\newblock Spatial as deep: Spatial cnn for traffic scene understanding.
\newblock In {\em Thirty-Second AAAI Conference on Artificial Intelligence},
  2018.

\bibitem{philion2019fastdraw}
Jonah Philion.
\newblock Fastdraw: Addressing the long tail of lane detection by adapting a
  sequential prediction network.
\newblock In {\em Proceedings of the IEEE/CVF Conference on Computer Vision and
  Pattern Recognition}, pages 11582--11591, 2019.

\bibitem{qin2020ultra}
Zequn Qin, Huanyu Wang, and Xi Li.
\newblock Ultra fast structure-aware deep lane detection.
\newblock In {\em Computer Vision--ECCV 2020: 16th European Conference,
  Glasgow, UK, August 23--28, 2020, Proceedings, Part XXIV 16}, pages 276--291.
  Springer, 2020.

\bibitem{qu2021focus}
Zhan Qu, Huan Jin, Yang Zhou, Zhen Yang, and Wei Zhang.
\newblock Focus on local: Detecting lane marker from bottom up via key point.
\newblock In {\em Proceedings of the IEEE/CVF Conference on Computer Vision and
  Pattern Recognition}, pages 14122--14130, 2021.

\bibitem{rezatofighi2019generalized}
Hamid Rezatofighi, Nathan Tsoi, JunYoung Gwak, Amir Sadeghian, Ian Reid, and
  Silvio Savarese.
\newblock Generalized intersection over union: A metric and a loss for bounding
  box regression.
\newblock In {\em Proceedings of the IEEE/CVF Conference on Computer Vision and
  Pattern Recognition}, pages 658--666, 2019.

\bibitem{su2021structure}
Jinming Su, Chao Chen, Ke Zhang, Junfeng Luo, Xiaoming Wei, and Xiaolin Wei.
\newblock Structure guided lane detection.
\newblock {\em arXiv preprint arXiv:2105.05403}, 2021.

\bibitem{sparsercnn}
Peize Sun, Rufeng Zhang, Yi Jiang, Tao Kong, Chenfeng Xu, Wei Zhan, Masayoshi
  Tomizuka, Lei Li, Zehuan Yuan, Changhu Wang, et~al.
\newblock Sparse r-cnn: End-to-end object detection with learnable proposals.
\newblock In {\em Proceedings of the IEEE/CVF Conference on Computer Vision and
  Pattern Recognition}, pages 14454--14463, 2021.

\bibitem{tabelini2021keep}
Lucas Tabelini, Rodrigo Berriel, Thiago~M Paixao, Claudine Badue, Alberto~F
  De~Souza, and Thiago Oliveira-Santos.
\newblock Keep your eyes on the lane: Real-time attention-guided lane
  detection.
\newblock In {\em Proceedings of the IEEE/CVF Conference on Computer Vision and
  Pattern Recognition}, pages 294--302, 2021.

\bibitem{tabelini2021polylanenet}
Lucas Tabelini, Rodrigo Berriel, Thiago~M Paixao, Claudine Badue, Alberto~F
  De~Souza, and Thiago Oliveira-Santos.
\newblock Polylanenet: Lane estimation via deep polynomial regression.
\newblock In {\em 2020 25th International Conference on Pattern Recognition
  (ICPR)}, pages 6150--6156. IEEE, 2021.

\bibitem{tusimple}
TuSimple.
\newblock Tusimple benchmark.
\newblock \url{https://github.com/TuSimple/tusimple-benchmark/}, Accessed
  September, 2020.

\bibitem{wang2018non}
Xiaolong Wang, Ross Girshick, Abhinav Gupta, and Kaiming He.
\newblock Non-local neural networks.
\newblock In {\em Proceedings of the IEEE conference on computer vision and
  pattern recognition}, pages 7794--7803, 2018.

\bibitem{xu2020curvelane}
Hang Xu, Shaoju Wang, Xinyue Cai, Wei Zhang, Xiaodan Liang, and Zhenguo Li.
\newblock Curvelane-nas: Unifying lane-sensitive architecture search and
  adaptive point blending.
\newblock In {\em Computer Vision--ECCV 2020: 16th European Conference,
  Glasgow, UK, August 23--28, 2020, Proceedings, Part XV 16}, pages 689--704.
  Springer, 2020.

\bibitem{yoo2020end}
Seungwoo Yoo, Hee~Seok Lee, Heesoo Myeong, Sungrack Yun, Hyoungwoo Park,
  Janghoon Cho, and Duck~Hoon Kim.
\newblock End-to-end lane marker detection via row-wise classification.
\newblock In {\em Proceedings of the IEEE/CVF Conference on Computer Vision and
  Pattern Recognition Workshops}, pages 1006--1007, 2020.

\bibitem{yu2018deep}
Fisher Yu, Dequan Wang, Evan Shelhamer, and Trevor Darrell.
\newblock Deep layer aggregation.
\newblock In {\em Proceedings of the IEEE conference on computer vision and
  pattern recognition}, pages 2403--2412, 2018.

\bibitem{yu2016unitbox}
Jiahui Yu, Yuning Jiang, Zhangyang Wang, Zhimin Cao, and Thomas Huang.
\newblock Unitbox: An advanced object detection network.
\newblock In {\em Proceedings of the 24th ACM international conference on
  Multimedia}, pages 516--520, 2016.

\bibitem{zhao2017pyramid}
Hengshuang Zhao, Jianping Shi, Xiaojuan Qi, Xiaogang Wang, and Jiaya Jia.
\newblock Pyramid scene parsing network.
\newblock In {\em Proceedings of the IEEE conference on computer vision and
  pattern recognition}, pages 2881--2890, 2017.

\bibitem{zheng2021resa}
Tu Zheng, Hao Fang, Yi Zhang, Wenjian Tang, Zheng Yang, Haifeng Liu, and Deng
  Cai.
\newblock Resa: Recurrent feature-shift aggregator for lane detection.
\newblock In {\em Proceedings of the AAAI Conference on Artificial
  Intelligence}, volume~35, pages 3547--3554, 2021.

\bibitem{zheng2021scaloss}
Tu Zheng, Shuai Zhao, Yang Liu, Zili Liu, and Deng Cai.
\newblock Scaloss: Side and corner aligned loss for bounding box regression.
\newblock {\em arXiv preprint arXiv:2104.00462}, 2021.

\end{thebibliography}
}

\end{document}